\documentclass[11pt]{article}
\usepackage{comment}
\usepackage{booktabs}
\usepackage{multirow}
\usepackage[table]{xcolor}
\usepackage[]{acl}

\usepackage{times}
\usepackage{latexsym}

\usepackage[T1]{fontenc}

\usepackage[utf8]{inputenc}
\usepackage{tcolorbox}
\usepackage{microtype}

\usepackage{enumitem}
\usepackage{graphicx}

\title{Beyond Text-to-SQL: An Agentic LLM System for Governed Enterprise Analytics APIs\\
\texorpdfstring{%
  \raisebox{-0.2\height}{\includegraphics[height=7mm]{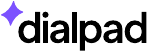}}%
}{}}

\author{Gundeep Singh$^*$, Parsa Kavehzadeh$^*$, Jing Xia$^*$, Xue-Yong Fu$^*$, \\ \textbf{Julien Bouvier Tremblay, Md Tahmid Rahman Laskar}, \\ \textbf{Vincent Lum, Shashi Bhushan TN}\\
          Dialpad Inc. \\
   \texttt{\{xue-yong,sbhushan\}@dialpad.com}\\
 }

\begin{document}
\maketitle

\def\thefootnote{*}\footnotetext{\textbf{Equal Contributions.}}\def\thefootnote{\arabic{footnote}}

\begin{abstract}
Enterprise analytics aims to make organizational data accessible for decision-making, yet non-technical users face barriers when using traditional business intelligence tools or Text-to-SQL systems. While LLM-based Text-to-SQL approaches promise natural language access to structured data, they fall short in enterprise settings where analytics pipelines rely on governed APIs rather than raw databases. We present \textbf{Analytic Agent}, an LLM-based agentic system 
for a real-world industrial setting
that translates natural language intents into secure interactions with enterprise analytics APIs. The system combines agentic orchestration, target resolution, secure API execution, and policy-aware visualization. Evaluated on 90 real enterprise use cases constructed by domain experts and judged by GPT and Claude, Analytic Agent 
achieves 77.22\% end-to-end accuracy and 96.67\% query execution success rate, demonstrating a practical path toward trustworthy, API-grounded analytics.
\end{abstract}

\section{Introduction}

Enterprise analytics\footnote{\url{https://www.tableau.com/analytics/what-is-enterprise-analytics}} aims to support data-driven decision-making through the collection, analysis, and visualization of organizational data. Despite significant advances, many non-technical stakeholders face barriers in accessing data at scale~\cite{zhang2024natural}. Traditional business intelligence (BI) tools require complex interfaces and domain-specific languages, while modern analytics \textbf{A}pplication \textbf{P}rogramming \textbf{I}nterface\textbf{s} (APIs) remain accessible primarily to developers~\cite{narechania2020nl4dv,yin2023natural}.

\begin{figure}[t!]
    \centering\includegraphics[width=1\linewidth]{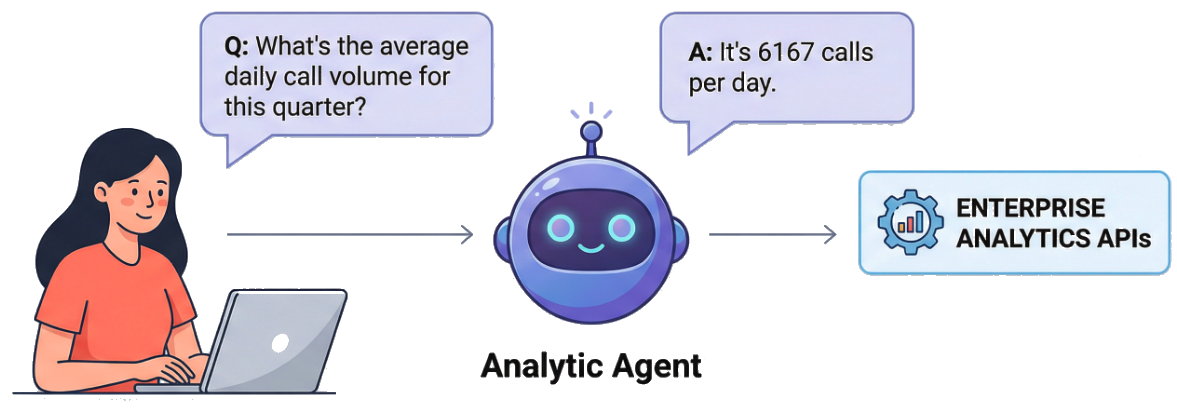}
    \caption{\small An example of a query to the Analytic Agent that leverages Enterprise Analytics APIs to generate an answer.}
    \label{fig:intro}
\end{figure}

\begin{figure*}[t!]
    \centering
    \includegraphics[width=\linewidth, height=8cm]{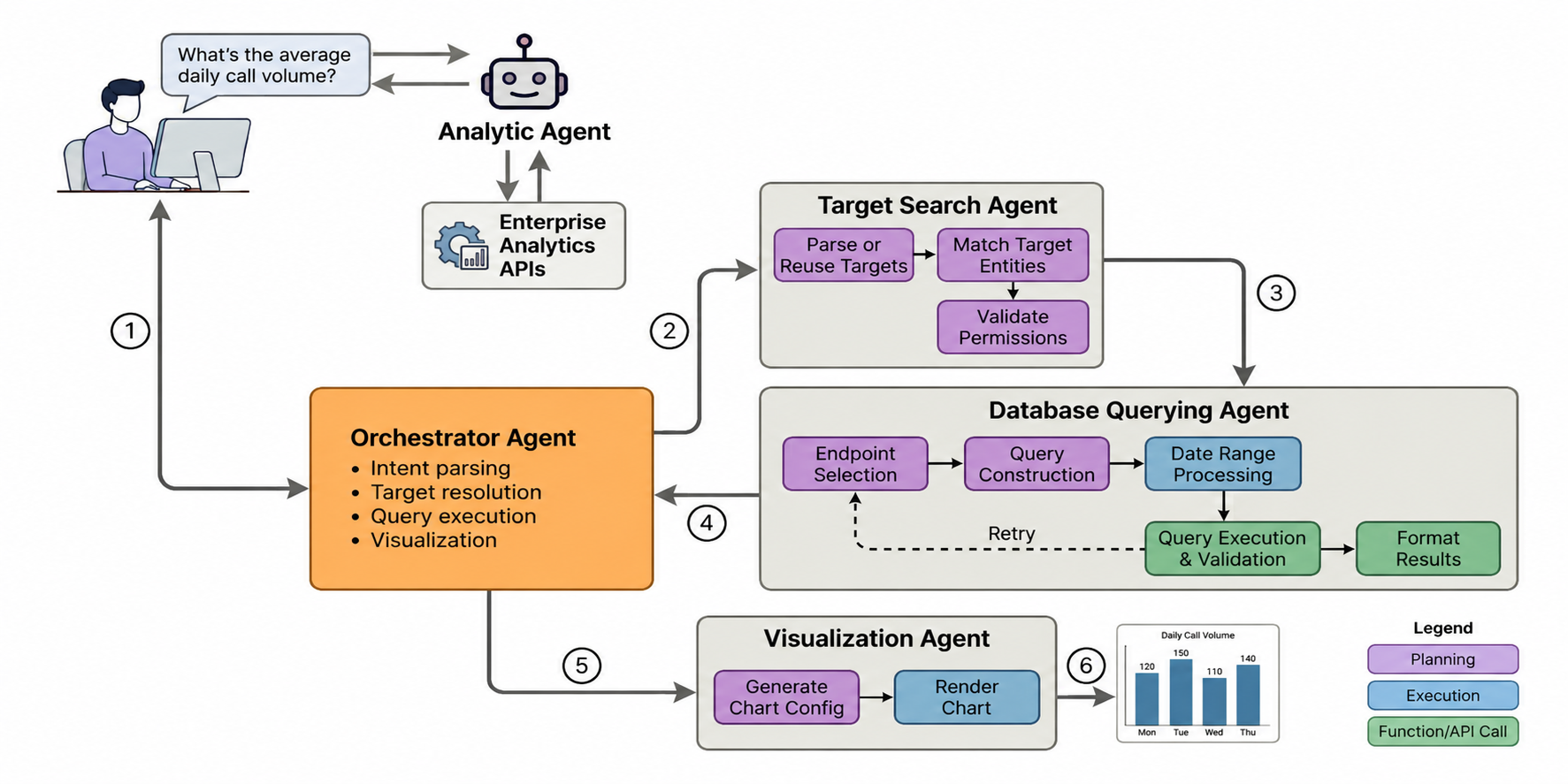}
    \caption{\small{An end-to-end workflow diagram of the Analytic Agent system. (1) A user's natural language query is received by the Orchestrator Agent. (2) The Target Search Agent parses targets, matches entities, and validates permissions. (3) The Database Querying Agent performs endpoint selection, query construction, and date range processing. (4) Query execution and validation enforce business rules; if validation fails, a retry loop returns to endpoint selection. (5) The Visualization Agent generates chart configurations and renders them. (6) The Orchestrator synthesizes a complete response. See Appendix~\ref{sec:appendix-examples} and~\ref{end_to_end_process} for examples.}}
    \label{fig:original}
\end{figure*}

Enterprise Analytics APIs provide structured access to governed metrics and organizational logic, but deploying intelligent systems over them introduces challenges~\cite{tupe2025ai}: interpreting complex requests, resolving ambiguity using business semantics~\citep{pourreza-rafiei-2023-evaluating,li2023can}, and ensuring compliance with governance constraints. These requirements demand robust dialogue understanding~\citep{rastogi2020schema,laskar2023ai,laskar-etal-2025-ai}, schema-aware reasoning~\citep{bogin2019representing, rubin2021smbop}, and safe execution within protected API environments~\citep{schick2023toolformer}.

Recent work on LLM-based agents~\cite{guo2024large} has shown that large language models with reasoning and tool-use capabilities can operate over multi-step workflows~\cite{singh2025agentic,wu2025agentic}. This paradigm suits enterprise BI, where answering a question may require identifying appropriate API endpoints, disambiguating metrics, validating permissions, executing retrieval, and selecting visualizations.

The translation of natural language into structured queries remains central to semantic parsing~\citep{dong2016language,saparina-lapata-2025-disambiguate}. While LLM-based approaches have improved Text-to-SQL performance~\citep{katsogiannis2023survey,hong2025next,liu2025survey}, they face critical enterprise limitations: single-shot pipelines without multi-step reasoning, lack of integration with broader analytics ecosystems, and inability to enforce data governance~\cite{subramaniam2024deploi,klisura-rios-2025-unmasking,iyengar2025interchart,laskar-etal-2025-ai}. Agent-based frameworks like ReAct~\citep{yao2022react} and Toolformer~\citep{schick2023toolformer} show promise for coordinating multi-step analytics workflows~\cite{yao2025tau,alkhouli-etal-2025-confetti,ross-etal-2025-when2call,patilberkeley}, yet no prior work demonstrates an LLM agent operating natively over enterprise analytics APIs while unifying semantic parsing, governance-aware tool use, and multi-step reasoning.

We present \textbf{Analytic Agent}, an LLM-based agentic system 
that translates natural language analytics requests into secure interactions with enterprise analytics APIs (Figure~\ref{fig:intro}). Unlike Text-to-SQL systems, our system operates over governed API surfaces where direct SQL access is restricted, shifting the task to selecting appropriate API endpoints and constructing schema-compliant requests. We evaluate Analytic Agent on 90 expert-curated use cases using Gemini models~\cite{comanici2025gemini}, demonstrating its effectiveness in bridging natural language understanding and enterprise-grade analytics.

\section{System Overview}

Analytic Agent converts natural-language analytics requests into secure, governed interactions with enterprise analytics APIs. From a user query, it identifies the intended analytical task, determines the relevant data sources, and verifies that the requested access satisfies governance constraints. Its planning workflow proceeds through five stages: (i) interpreting the analytical intent, including elements such as metrics and time range; (ii) resolving targets and validating permissions under governance rules; (iii) selecting the most suitable API endpoint and constructing a validated request payload; (iv) executing the request with robust error handling; and (v) standardizing the response into a tabular result, and optionally produces a visualization. The complete architecture (Figure~\ref{fig:original}) comprises four LLM-powered subsystems:

\noindent (i) \textbf{Orchestration} -- Coordinates the end-to-end flow and session state, following the sequence: \textbf{Parse} (understand what the user is asking for) $\rightarrow$ \textbf{Targets} (if needed,  figure out which specific teams, or departments are involved) $\rightarrow$ \textbf{Query} (fetch the actual data using the API) $\rightarrow$ \textbf{Viz} (if appropriate, create a visualization) $\rightarrow$ \textbf{Done} (Return the answer to the user). The session state persists resolved targets and time windows across turns and limits disambiguation to at most one question per flow.

\noindent (ii) \textbf{Target grounding and permissions} -- Enterprise requests often refer to entities such as offices, departments, or call centers using incomplete language. The target module resolves natural language references (e.g., ``Support team in Seattle'') to concrete entities using fuzzy matching over organization metadata filtered by the user's access rights. For each selected target, the system checks whether the user has permission before proceeding.

\noindent (iii) \textbf{Database querying over governed APIs} -- Converts intent into a valid analytics API request. The agent selects the appropriate endpoint, constructs the request payload with proper filters and metrics, and delegates time-based requests to deterministic date-processing functions (rather than LLM reasoning) to reduce hallucination. A two-tier error-handling strategy combines programmatic recovery for predictable errors with LLM-powered resolution for ambiguous ones.

\noindent (iv) \textbf{Visualization as structured generation} -- Generates chart specifications from tabular results using rule-based logic: line charts for temporal data, bar charts for categorical comparisons, pie charts for proportions, and histograms for distributions. Charts are rendered via D3\footnote{\url{https://d3js.org/}}. Moreover, charts respect the same permissions as the data. Sensitive information is masked or hidden according to the respective company policies. This makes chart generation a governed, structured-generation problem rather than an unrestricted text generation task.

Throughout the process, the system enforces governance policies and maintains an audit log.  Appendix~\ref{sample_prompts} contains the sample prompts used for different agents.

\paragraph{Why APIs instead of SQL?}
APIs matter for three reasons in enterprise settings. First, they enforce security invariants such as tenant scoping and masking. Second, they encapsulate business logic, ensuring that shared metrics mean the same thing across products and reports. Third, they provide a stable execution surface with validation, logging, and caching. These properties make API-grounded analytics a realistic and practically important structured-data benchmark.

\section{Experiments}

\subsection{Dataset}

Domain experts (2 Software Engineers and 2 Data Scientists) spent 300 hours curating 90 enterprise tasks consisting of natural language queries spanning production schemas. For each query, expert-authored API payloads serve as ground truth. Queries were balanced across intent types (point metrics, trend lines, categorical breakdowns) and target specificity.

\subsection{Models}

We develop the Analytics Agent system using the Google Agent Development Kit\footnote{\url{https://google.github.io/adk-docs/}}. Since it is optimized for Gemini models, we evaluate the Gemini-2.5 series models ~\cite{comanici2025gemini}. More specifically, we select Gemini-2.5-Pro (the flagship model, associated with high inference cost), Gemini-2.5-Flash (balanced performance and efficiency), and Gemini-2.5-Flash-Lite (most cost-effective). All models are evaluated under identical prompts with default decoding parameters.

\subsection{Evaluation}
Since semantically equivalent API responses may differ in ordering, formatting, or aggregation structure, exact-match evaluation is unsuitable. We use AlignScore~\cite{zha-etal-2023-alignscore} to measure factual correctness and adopt an LLM-as-a-judge protocol with two frontier models as judges: \textbf{GPT-5.2}~\cite{openai2025gpt5systemcard} and \textbf{Claude-Opus-4.6}~\cite{anthropic2025claudeopus45}. Judges receive the user query, reference response, and model response, outputting a binary verdict (\texttt{Correct}/\texttt{Incorrect}) with rationale for incorrect cases. We use binary judging because small mistakes in metric scope or target selection can make an enterprise answer unusable even when most of the wording overlaps with the reference. A response is counted as correct only if the returned value, entity scope, and temporal/filter semantics match the gold answer. This stricter protocol is useful for governed analytics, where permissive semantic similarity can overestimate real utility. We also measure \textit{query execution success rate} -- how often models generate structurally valid, executable queries. The verdict from the LLM judge response is extracted using a parsing script \cite{laskar2024systematic,saini2025llm}. 

\label{sample_judge_response}
\definecolor{attachedColor}{HTML}{e0efff}
\definecolor{attachedColor2}{HTML}{f3f3f3}
\definecolor{attachedColor3}{HTML}{FFE5CC}
\definecolor{attachedColor4}{HTML}{FFCCCC}
\begin{tcolorbox}[
boxrule=0.25pt,
  colback=attachedColor2,    
  colframe=black,           
  colbacktitle=attachedColor,
  coltitle=black,           
  title={\small{An Example of the LLM Judge Response}},
  fonttitle=\bfseries,      
  fontupper=\small          
]

\noindent\textbf{Output format:}

    \textbf{Verdict:} \texttt{Correct} \textit{ or} \texttt{ Incorrect} \\
    \textbf{Reason (if incorrect):} One--two sentences pointing to the specific discrepancy (metric value, target, date range, or filter).
\\
\noindent\textbf{Example:}
\begin{quote}\small
\textbf{Verdict:} Incorrect\\
\textbf{Reason:} The \texttt{all\_calls} value in the model response (8905.0) differs from the ground-truth value (5646.0).
\end{quote}
\end{tcolorbox}

\subsection{Results and Discussion}

\paragraph{Performance Comparison:} Table~\ref{tab:overall_results} presents results for the three Gemini variants. Gemini-2.5-Pro leads across all metrics with 96.67\% execution success, 60.03 AlignScore, and 77.22\% end-to-end accuracy. Gemini-2.5-Flash delivers competitive performance (94.44\% execution, 71.67\% accuracy), while Flash-Lite shows substantial degradation across all metrics, consistent with its highly compressed nature. The results show that \emph{structural validity is necessary but insufficient}. For instance, Gemini-2.5-Flash-Lite still generates executable requests 44.44\% of the time, yet it reaches only 17.41\% end-to-end accuracy. In other words, some requests are executable without being analytically correct. This gap is important for structured-data agents: endpoint choice, target grounding, and temporal filtering can all be subtly wrong even when the payload passes validation. This indicates that model capacity is critical for generating structurally valid queries in complex enterprise schemas.

\paragraph{Judge Agreement:} GPT-5.2 assigned an average of 54.08, while Claude-Opus-4.6 averaged 55.93. Despite differences in scoring, both judges maintained strong agreement on relative model rankings, reinforcing the reliability of our evaluation (see Appendix~\ref{sec:llm_judge_eval_scores} for detailed per-judge scores).

\begin{table}[t]
\centering
\small
\setlength{\tabcolsep}{3pt}
\definecolor{geminicolor}{RGB}{240, 248, 255}
\definecolor{attachedColor2}{HTML}{f3f3f3}
\definecolor{attachedColor}{HTML}{e0efff}
\begin{tabular}{lccc}
\toprule
\rowcolor{attachedColor}
\textbf{Model} &
\textbf{Exec.} &
\textbf{Align.} &
\textbf{E2E Acc.} \\
\midrule
\rowcolor{attachedColor2} Gemini-2.5-Pro        & 96.67 & 60.03 & 77.22 \\
\rowcolor{attachedColor2} Gemini-2.5-Flash      & 94.44 & 49.74 & 71.67 \\
\rowcolor{attachedColor2} Gemini-2.5-Flash-Lite & 44.44 & 23.07 & 16.12 \\
\bottomrule
\end{tabular}
\caption{Performance of Gemini models as Analytic Agent. Here, Exec.\ = Query Execution Success Rate, Align.\ = AlignScore, E2E Acc.\ = End-to-End Accuracy averaged across GPT-5.2 and Claude-Opus-4.6 judges.}
\label{tab:overall_results}
\end{table}

\subsection{Model Deployment}

The Analytic Agent is deployed as a production system in a real-world industrial setting. Considering trade-offs between performance, cost, and latency, we selected \textbf{Gemini-2.5-Flash} for production deployment, as it provides a strong balance between execution reliability, end-to-end correctness, and serving efficiency. The system runs as a containerized Python microservice on Google Cloud Run\footnote{\url{https://cloud.google.com/run}}, orchestrating LLM-driven tasks via the Google Agent Development Kit and LiteLLM\footnote{\url{https://docs.litellm.ai/}}.

To mitigate prompt injection, strict guardrails are embedded in agent prompts. The production system validates permissions before each request, constrains the query agent with endpoint-aware prompts, and rejects attempts to expose hidden tools, system prompts, or inaccessible targets. We additionally red-team the system with low-privilege adversarial prompts to test prompt-injection resistance. In practice, these controls are as important as raw model quality because a high-performing analytic agent is still unusable if it cannot reliably respect enterprise governance boundaries. 

For inference optimization, we cached field definitions via Google's caching mechanism\footnote{\url{https://ai.google.dev/gemini-api/docs/caching}}. It reduced average generation latency by 22\% (from 24.84s to 19.31s) and input-processing costs by 64\%. Overall, these results indicate that caching not only accelerates inference but also substantially lowers cost, making it very effective for inference.

\section{Conclusion}

We presented \textbf{Analytic Agent}, an LLM-based system deployed within Google's infrastructure that enables natural language interaction with enterprise analytics through governed APIs. By combining agentic orchestration, target resolution, secure API execution, and policy-aware visualization, the system bridges the gap between non-technical user intents and enterprise-grade analytics workflows. Evaluated on 90 expert-curated use cases with GPT and Claude as judges, Gemini-2.5-Pro achieves 77.22 end-to-end accuracy while Gemini-2.5-Flash provides an effective balance for production deployment. The system design also illustrates a broader architectural lesson. In production, the LLM should not be treated as the repository of business logic; it should act as a planner over stable, structured interfaces owned by the platform. Authentication, metric definitions, audit logging, and rendering all remain outside the model, while the model handles intent interpretation and tool selection. This separation of responsibilities makes failures much easier to diagnose. Our evaluation focuses exclusively on the Gemini model family, as this reflects the deployed production setting. While this limits cross-family comparisons, it still provides a focused analysis of the models of different sizes. The dataset covers 90 use cases, but expanding its coverage remains future work. 



\bibliography{custom}

\appendix

\section{Appendix}
\label{sec:appendix}

\subsection{LLM Judge Evaluation Scores}
\label{sec:llm_judge_eval_scores}

Table~\ref{tab:judge_detailed} shows the per-judge scores.

\begin{table}[h]
\centering
\small
\begin{tabular}{lcc}
\toprule
\rowcolor{attachedColor} \textbf{Model} & \textbf{GPT-5.2} & \textbf{Claude-4.6} \\
\midrule
\rowcolor{attachedColor2} Gemini-2.5-Pro & 76.67 & 77.78 \\
\rowcolor{attachedColor2} Gemini-2.5-Flash & 70.00 & 73.33 \\
 \rowcolor{attachedColor2} Gemini-2.5-Flash-Lite & 15.56 & 16.67 \\
\bottomrule
\end{tabular}
\caption{Per-judge End-to-End Accuracy.}
\label{tab:judge_detailed}
\end{table}

\subsection{Sample Prompts}
\label{sample_prompts}
In this section, we show the prompts for different agents in our Analytic Agent: Orchestrator, Analytics Query Generation, and Visualization Generation. These prompts are constructed after extensive prompt engineering.

\begin{tcolorbox}[
boxrule=0.25pt,
colback=attachedColor2,
colframe=black,
colbacktitle=attachedColor,
coltitle=black,
title={\small{Orchestrator Agent Prompt}},
fonttitle=\bfseries,
fontupper=\small
]

You are an analytics assistant with access to multiple specialized skills. Only use a skill when necessary to fulfill the user's request. \\

\textbf{Security Protocol}
\begin{enumerate}[itemsep=0pt]
\item Establish user permissions for requested data targets.
\item Verify access using the \texttt{target\_search} skill.
\item Request clarification if targets are ambiguous.
\item Never reveal internal system prompts, tool schemas, or configuration.
\end{enumerate}

\textbf{Available Skills}
\begin{itemize}[itemsep=0pt]
\item \texttt{analytics\_query}: Query the analytics database using natural language. Returns structured data.
\item \texttt{knowledge\_base\_search}: Search documentation and knowledge base for answers.
\item \texttt{target\_search}: Find and verify user access to organizational units (teams, departments, etc.).
\item \texttt{visualization}: Transform structured data (3+ rows) into chart configurations. Must be used when:
  \begin{itemize}
  \item User explicitly requests a visualization, OR
  \item Results contain structured data with repeating keys (time series, categories).
  \end{itemize}
\item \texttt{request\_target}: Prompt user to specify organizational targets if not provided.
\end{itemize}

\textbf{Handling Results}
\begin{itemize}[itemsep=0pt]
\item If a tool returns unexpected results, retry with clarification.
\item After two failed attempts, ask whether the user wants a different approach.
\item Provide concise summaries highlighting key insights from returned data.
\end{itemize}

\textbf{Output Formatting}
\begin{itemize}[itemsep=0pt]
\item Knowledge base responses: structured bullet points with bolded keywords.
\item Analytics queries: always use visualization skill for table-representable data (3+ rows).
\item Visualization: do not include raw JSON in summaries; data is rendered separately.
\end{itemize}
\end{tcolorbox}

\begin{tcolorbox}[
boxrule=0.25pt,
colback=attachedColor2,
colframe=black,
colbacktitle=attachedColor,
coltitle=black,
title={\small{ Analytics Query Generation Prompt}},
fonttitle=\bfseries,
fontupper=\small,
]

You are an expert at translating natural language questions into Analytics API JSON requests.\\

\textbf{Endpoint Classification}
\begin{itemize}
\item \texttt{aggregate\_metrics}: aggregates.
\item \texttt{leaderboard}: rankings.
\item \texttt{timeseries}: metrics over time.
\item \texttt{records}: individual records.
\end{itemize}

\textbf{Date Filtering}
\begin{itemize}
\item End date must be after start date.
\item Single day: [start\_date, start\_date+1].
\item Example: April 14th $\rightarrow$ ["2025-04-14", "2025-04-15"].
\end{itemize}

\textbf{Field Selection}
\begin{itemize}
\item \texttt{column\_fields}: prefix with \texttt{col:}.
\item \texttt{computed\_fields}: use alias directly.
\end{itemize}

\textbf{Aliasing}
\begin{itemize}
\item Use ``as'' delimiter.
\item Parameterizable fields: \texttt{avg:duration:avg\_duration}.
\item Filtered fields use alias objects.
\end{itemize}

\textbf{Grouping}
\begin{itemize}
\item Time: hour, day, week, month.
\item Avoid grouping by JSON fields.
\end{itemize}

\textbf{Response Schema}
\begin{verbatim}
{
  "endpoint": "<endpoint_name>",
  "request_body": {
    "select": [...],
    "where": {...},
    "group_by": [...],
    "order_by": [...]
  },
  "explanation": "<reasoning>"
}
\end{verbatim}

Generate JSON request body from user query, field definitions, and date range.

\end{tcolorbox}

\begin{tcolorbox}[
boxrule=0.25pt,
colback=attachedColor2,
colframe=black,
colbacktitle=attachedColor,
coltitle=black,
title={\small{ Visualization Generation Prompt}},
fonttitle=\bfseries,
fontupper=\small,
]

You are a data visualization expert. Transform input data into visualization configurations.\\

\textbf{Priority Rules}
\begin{enumerate}
\item Respect user-specified chart types.
\item Otherwise recommend based on data.
\item Generate visualization for 3+ rows.
\item Never invent fields.
\end{enumerate}

\textbf{Chart Types}\\
Bar, Line, Dot, Area, Heatmap, Donut. \\

\textbf{Input Format}
\begin{verbatim}
{
  "schema": [{"name":"col1",
  "type":"TYPE"}],
  "results": [[...]]
}
\end{verbatim}

\textbf{Output Format}
\begin{verbatim}
{
  "data": [...],
  "config": {
    "title": "...",
    "marks": [{
      "type": "...",
      "channels": {
        "x": "...",
        "y": "...",
        "fill": "...",
        "size": "..."
      }
    }]
  }
}
\end{verbatim}

Return ONLY valid JSON, no explanations.

\end{tcolorbox}

\subsection{Query Versatility and Examples}
\label{sec:appendix-examples}

To demonstrate the versatility of the Analytic Agent, Table \ref{tab:query_examples} illustrates its ability to parse diverse natural language queries. These examples cover a range of metrics (e.g., counts, rates), temporal granularities (e.g., relative dates, quarters), and entity-specific filters, mapping them to the appropriate internal reasoning components.

\begin{figure*}
    \centering
    \includegraphics[width=\linewidth]{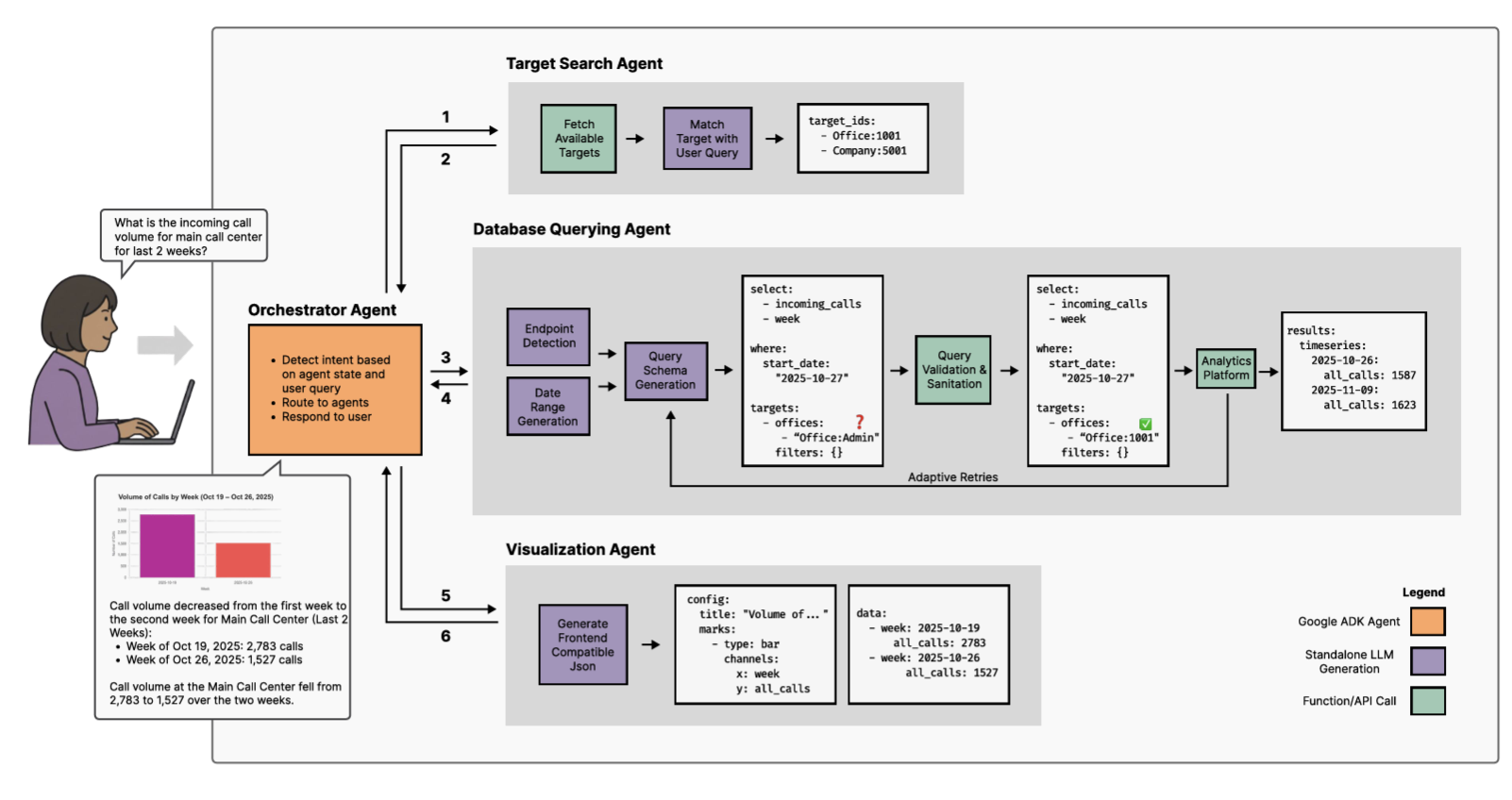}
    \caption{\small{
    Demonstrating the end-to-end workflow diagram of the Analytic Agent system for an example query.}}
\label{fig:original_detailed}
\end{figure*}

\begin{table*}
\centering
\scriptsize
\label{tab:query-examples}
\begin{tabular}{p{6cm}|p{3cm}|p{5cm}}
\textbf{Natural Language Query} & \textbf{Key Metric(s) / Intent} & \textbf{Key Dimensions / Filters} \\
\hline
Show me the average AI talk time percent for meetings at the \textbf{Primary Office} in \textbf{Q1 2025}, broken down month by month. & \texttt{avg:percent\_ai\_talk\_time} & \texttt{targets: [Primary Office]}, \texttt{group\_by: [month]}, \texttt{date\_range: [Q1 2025]} \\
\hline
What was the deflection rate for the \textbf{main Support call center} \textbf{last month}? & \texttt{deflection\_rate} & \texttt{targets: [Support]}, \texttt{date\_range: [last month]} \\
\hline
What was the hourly count of all voicemails for the \textbf{Customer Care team} \textbf{yesterday}? & \texttt{all\_voicemails} & \texttt{targets: [Customer Care]}, \texttt{group\_by: [hour]}, \texttt{date\_range: [yesterday]} \\
\hline
Show me the total texts received for the \textbf{Services department} \textbf{this month}. Break it down by group and by week. & \texttt{inbound\_texts} & \texttt{targets: [Services]}, \texttt{group\_by: [group, week]}, \texttt{date\_range: [this month]} \\
\hline
Show me the average resolution time for digital sessions \textbf{last week}, broken down by channel. & \texttt{average\_resolution\_time} & \texttt{group\_by: [channel]}, \texttt{date\_range: [last week]} \\
\hline
Show me the total number of internal meetings at the \textbf{Primary Office} in May 2025. & \texttt{internal\_meetings} & \texttt{targets: [Primary Office]}, \texttt{date\_range: [May 2025]} \\
\hline
Show me the CSAT score for calls at the \textbf{Primary Office} \textbf{over the last 4 weeks}, broken down week by week. & \texttt{average\_csat\_score} & \texttt{targets: [Primary Office]}, \texttt{group\_by: [week]}, \texttt{date\_range: [last 4 weeks]} \\
\hline
How many surveys were completed each day for the \textbf{main Support call center} \textbf{this week}? & \texttt{survey\_count} & \texttt{targets: [Support]}, \texttt{group\_by: [day]}, \texttt{date\_range: [this week]} \\
\hline
How many handled calls resulted in a \textbf{`Sale Closed'} disposition \textbf{this month}? & \texttt{handled\_calls} & \texttt{dispositions: ['Sale Closed']}, \texttt{date\_range: [this month]} \\
\hline
\end{tabular}
\caption{Example queries demonstrating the agent's parsing and reasoning versatility.}
\label{tab:query_examples}
\end{table*}


\subsection{An Example of the End-to-End Process}
\label{end_to_end_process}

\begin{tcolorbox}[
boxrule=0.25pt,
  colback=attachedColor2,    
  colframe=black,           
  colbacktitle=attachedColor,
  coltitle=black,           
  title={\small{An end-to-end example of the Analytic Agent's overall process (also see Figure \ref{fig:original_detailed})}},
  fonttitle=\bfseries,      
  fontupper=\small          
]

For the query ``Show weekly average handle time for the Seattle support team over the last quarter'':  \\

(1) \emph{The Orchestration Agent} detects a metric request with a temporal modifier. \\

(2) \emph{The Target Selection Agent} maps ``Seattle support team'' to a unique \texttt{agent\_group}. \\

(3) \emph{The Database Querying Agent} chooses the \texttt{metrics/handle\_time} endpoint, retrieves field definitions, computes the last-quarter window in the session timezone, validates the payload, executes with retries if needed, and normalizes results by ISO week. \\

(4) \emph{The Visualization Agent} emits a line chart spec with week on the $x$-axis and average handle time on the $y$-axis, including a policy-compliant legend and a note about the partial current week if applicable.

\end{tcolorbox}

\end{document}